\documentclass[runningheads]{llncs}
\usepackage[T1]{fontenc}
\usepackage{graphicx,verbatim}
\usepackage{float}
\usepackage{amsfonts, amsmath, amssymb}
\usepackage{xcolor}
\begin{document}
\title{Context Matters: Support Set Selection and Failure Detection for In-Context Medical Image Segmentation}
%

\author{Youssef Gehad, Emmanuel Zerefa, Krish Kabra, Guha Balakrishnan}  
\authorrunning{Gehad et al.}
\institute{Rice University \\
    \email{yag1@rice.edu}\\
    \email{emz1@rice.edu} \\
    \email{kk80@rice.edu} \\
    \email{guha@rice.edu}}
  
\maketitle              

\begin{abstract}
In-context learning (ICL) adapts medical image segmentation models to unseen structures and modalities without retraining by conditioning on a task-specific support set of image-mask exemplars. Because this support set is the model's only task-specific signal, its composition directly influences segmentation performance. In this work, we investigate the support set as a controllable determinant of ICL reliability. First, we compare random sampling against similarity-based selection, where exemplars are retrieved based on their visual similarity to the query image. Second, we train a transformer-based classifier to predict, from the query and support images alone, whether a segmentation will fall below a specified Intersection-over-Union (IoU) threshold. Using MultiverSeg with DINOv3 embeddings across four benchmarks and three imaging modalities, we show that similarity-based selection consistently matches or outperforms random sampling, with the largest gains at the smallest support set sizes. Furthermore, our classifier predicts segmentation failure above chance on all four benchmarks. Ultimately, these results demonstrate that the reliability of in-context segmentation can be both improved via informed support selection and anticipated before use, providing practical mechanisms for safer clinical deployment.
\end{abstract}
\section{Introduction}

Image segmentation is fundamental to biomedical image analysis across modalities including MRI, CT, ultrasound, and histopathology, enabling critical clinical applications such as disease monitoring~\cite{de2018clinically,shi2025automated} and treatment planning~\cite{harrison2022machine}. Although deep learning has achieved expert-level accuracy across numerous segmentation tasks~\cite{greenwald2022whole,isensee2021nnu,wasserthal2023totalsegmentator}, conventional models remain rigidly task-specific. Adapting a trained model to a novel modality or anatomical structure requires explicit retraining or fine-tuning, a process that typically demands substantial compute and hundreds to thousands of dense annotations.

Recently, in-context learning (ICL) has emerged as a promising alternative~\cite{butoi2023universeg,rakic2024tyche,wong2025multiverseg,gao2025show}. Originally popularized in natural language processing, where models perform novel tasks by conditioning on a few ``prompt'' examples at inference time rather than updating their weights~\cite{brown2020language,dong2024survey}, ICL has now been adapted for image segmentation. By predicting a mask for a query image based on a support set of reference image–mask pairs, ICL models can segment unseen anatomies, modalities, and targets without retraining. This flexibility is especially compelling for medical imaging, where defining new tasks on the fly with just a handful of exemplars could substantially lower the barrier to clinical deployment.

However, the performance of in-context segmentation depends heavily on the composition of the support set, not just its size. As the model's sole task-specific signal, the support set alone dictates how a query is interpreted. Consequently, a poorly matched support set can severely degrade segmentation quality compared to a well-matched one, even when the set size is fixed~\cite{suo2024visual,zhang2023makes}. This sensitivity raises two critical questions for safe clinical deployment: (i) what constitutes an effective support set for a given query, and (ii) when will a specific query–support pairing yield an unreliable result?

In this work, we address both questions empirically. First, we evaluate how support selection strategies impact performance, comparing random sampling against a similarity-based approach where reference images are visually matched to the query~\cite{suo2024visual,sun2025exploring,zhang2023makes}. Second, we investigate whether segmentation failure can be anticipated directly from this context. Specifically, we train a classifier to predict whether a resulting segmentation will meet a target Intersection-over-Union (IoU) threshold using only the query and support image embeddings. Across four biomedical imaging benchmarks, we demonstrate that similarity-based selection consistently matches or outperforms random sampling, with the most significant gains occurring at the smallest support set sizes. Moreover, our classifier successfully predicts segmentation failure above chance on all tasks. These findings provide actionable mechanisms through informed selection and proactive failure detection to improve the reliability of in-context segmentation for clinical deployment.

\section{Methods}~\label{sec:methods}
We consider the standard in-context segmentation setting. Given a query image $x_q$ and a support set $\mathcal{S}=\{(x_s^i,y_s^i)\}_{i=1}^{K}$ of $K$ reference image–mask pairs, an in-context model $f$ predicts a binary segmentation mask $\hat{y}_q=f(x_q,\mathcal{S})$ for the target structure implicitly defined by $\mathcal{S}$. To represent images independently of the segmentation model, we use a frozen image encoder $g$ to map an image $x$ to an embedding $e=g(x)$. We measure segmentation quality using the Intersection-over-Union (IoU) between the predicted mask $\hat{y}_q$ and the ground-truth mask $y_q$.

Given an initial dataset $\mathbb{D}=\{(x^i,y^i)\}_{i=1}^{M}$ of $M$ image–mask pairs, we partition it once into a query pool $\mathbb{D}_q$ and a support pool $\mathbb{D}_s$ of sizes $M_q$ and $M_s$. Each support set $\mathcal{S}$ is sampled from $\mathbb{D}_s$, and every query $(x_q,y_q)$ is drawn from $\mathbb{D}_q$. This disjoint split ensures that no image ever serves as both a query and a support example.

\subsection{Support Set Selection}
~\label{sec:selection}

\begin{figure}[t]
    \centering
    \includegraphics[width=\textwidth]{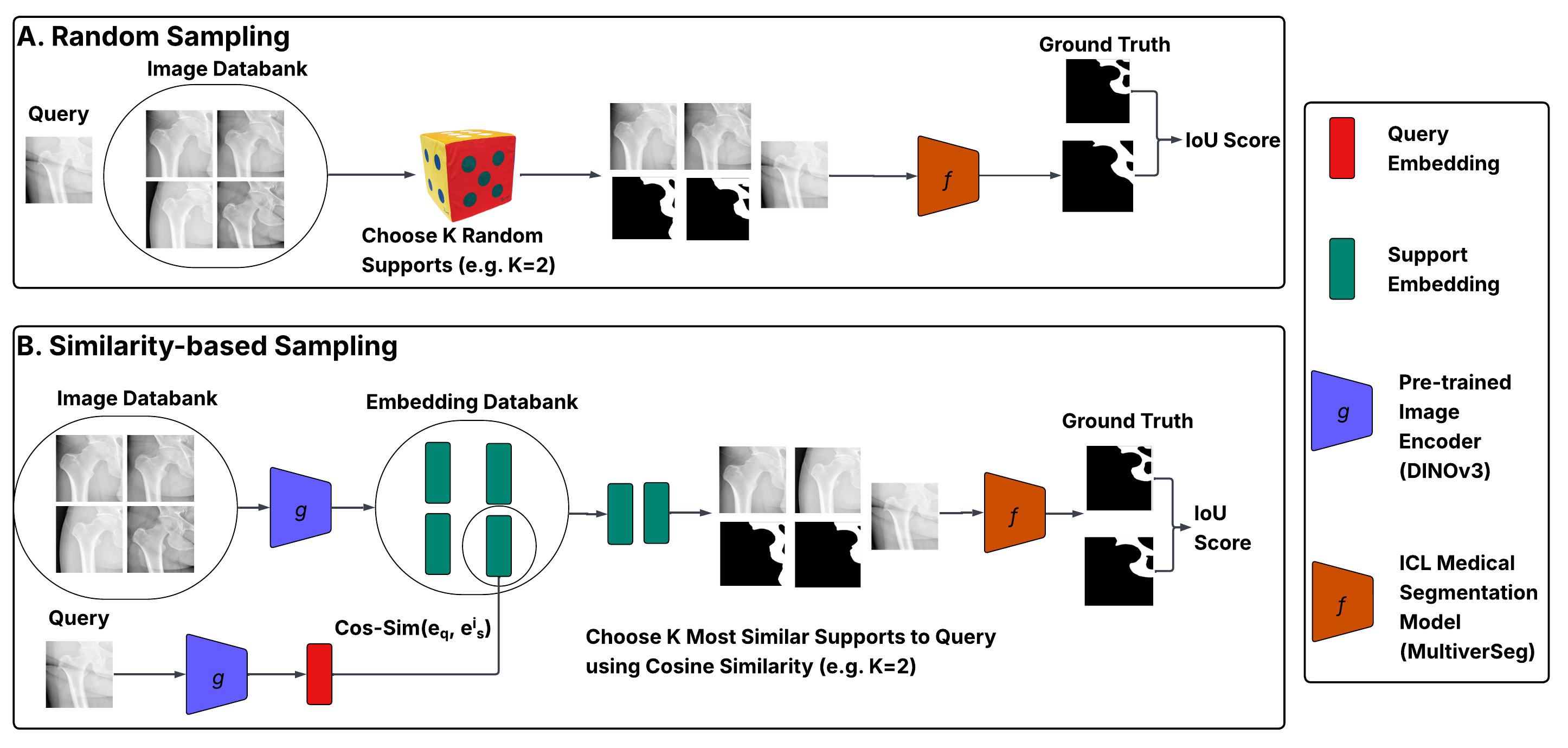}
    \caption{\textbf{Overview of evaluated support set sampling strategies.} \textbf{A.} Random sampling selects $K$ support images uniformly from the support dataset. \textbf{B.} Similarity-based sampling retrieves the $K$ support examples with the highest cosine similarity to the query embedding. For both strategies, the in-context model uses the query and the selected support set to predict a segmentation mask, with performance evaluated via the Intersection-over-Union (IoU) against the ground truth. Because random sampling is stochastic, we repeat this draw for $N=40$ independent trials per query and report the average IoU. Similarity-based sampling, on the other hand, is deterministic and therefore requires only a single sample per query.}
    \label{fig:sampling_pipeline}
\end{figure}

To study how support set composition affects performance, we compare two strategies for assembling the support set $\mathcal{S}$ for a given query: random sampling and similarity-based sampling. Figure~\ref{fig:sampling_pipeline} provides an overview of these strategies. 

\paragraph{Random sampling.} As a baseline, we consider the naive strategy in which the support set carries no query-specific information. We draw $K$ pairs uniformly at random from the support pool $\mathbb{D}_s$. Because this draw is stochastic, a single sample reflects only one of many possible support sets. We therefore repeat the procedure for $N=40$ independent trials per query and characterize performance using the average IoU across trials.

\paragraph{Similarity-based sampling.} A popular strategy to tailor the support set is to select reference images visually similar to the query~\cite{suo2024visual,sun2025exploring,zhang2023makes}. For a given query, we compute its embedding $e_q=g(x_q)$ and rank the support images by the cosine similarity between $e_q$ and each support embedding $e_s^i = g(x_s^i)$. We then define $\mathcal{S}$ as the $K$ pairs most similar to the query. Unlike random sampling, this selection is deterministic given the embeddings, requiring only a single trial per query.

\subsection{In-context Segmentation Failure Detection}
~\label{sec:prediction}

\begin{figure}[t]
    \centering
    \includegraphics[width=\textwidth]{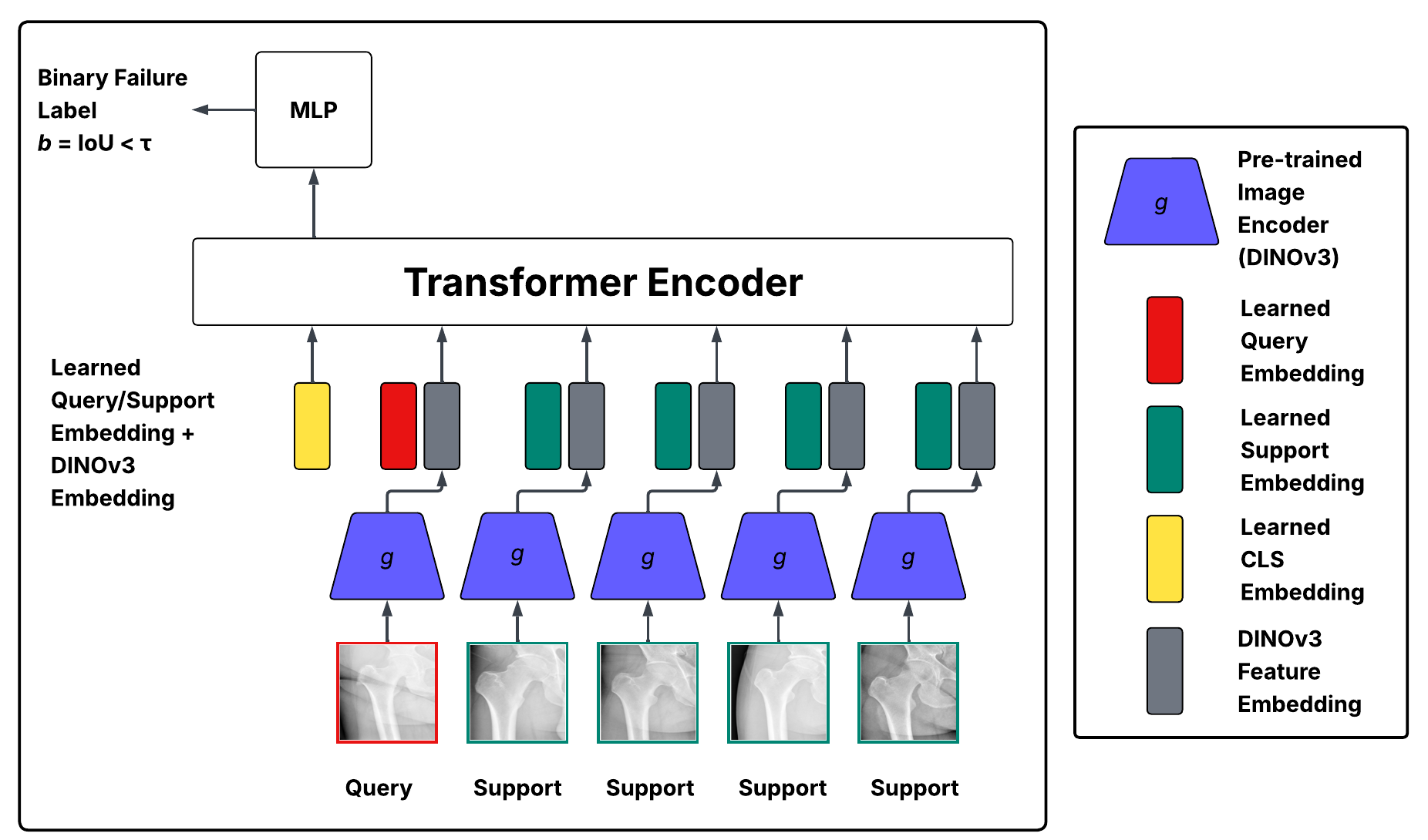}
    \caption{\textbf{Overview of the proposed in-context segmentation failure detection architecture.} We use a transformer-based classifier to predict segmentation failure, defined as a binary indicator of whether the resulting segmentation falls below a predefined IoU threshold $\tau$. First, a pre-trained image encoder (e.g., DINOv3) extracts embeddings for the query and support images. To distinguish between these roles, we add learned query and support embeddings to their respective image embeddings. These combined embeddings, alongside a learned class token (\textsc{[cls]}), serve as inputs to a standard transformer encoder. Finally, the \textsc{[cls]} output token passes through a classification head to predict the failure label. }
    \label{fig:detector_pipeline}
\end{figure}

To study whether in-context segmentation failure can be anticipated from context alone, without producing the mask, we cast the problem as a binary classification task: given a query and a support set, predict whether the resulting segmentation will fall below a pre-defined IoU threshold $\tau$. We solve this task in a supervised fashion, constructing a labeled dataset via the random sampling strategy (Section~\ref{sec:selection}) and training a transformer-based classifier on the query and support-set image embeddings. We describe each component below.

\paragraph{Labeled dataset construction.} To enable supervised learning, we generate a labeled dataset of query–support inputs $(x_q, \mathcal{S})$ using the random sampling procedure. We pair each input with a binary failure label $b$ indicating whether the resulting segmentation is of poor quality:

\begin{equation}
    b = \begin{cases}
        1 & \text{if } \mathrm{IoU}(\hat{y}_q, y_q) < \tau \\
        0 & \text{otherwise,}
    \end{cases}
\end{equation}
where we compute the IoU between the predicted mask $\hat{y}_q = f(x_q, \mathcal{S})$ and the ground truth $y_q$.

\paragraph{Transformer-based failure classifier.} We design a transformer-based classifier~\cite{dosovitskiy2021an,vaswani2017attention} that takes query and support-set image embeddings as input to predict the binary failure label $b$. Figure~\ref{fig:detector_pipeline} illustrates an overview of the model. Given an input $(x_q, \mathcal{S})$, we use the corresponding query and support image embeddings, $e_q$ and $\{e_s^i\}_{i=1}^{K}$ respectively, as input tokens to the transformer. To help the model distinguish between the two roles, we add a learned query embedding to the query token and a learned support embedding to each support token. We apply no positional embeddings among the support tokens; this ensures the classifier remains invariant to the ordering of the support set, consistent with the set-valued nature of $\mathcal{S}$. Finally, we introduce a learned class token, \textsc{[cls]}, whose output passes through a classification head to predict the failure label $\hat{b}$.

\section{Experiments and Results}
\label{sec:experiments}

\subsection{Experimental Setup}
\label{sec:setup}
\paragraph{Datasets.} We conducted our experiments on four biomedical image segmentation benchmarks spanning diverse imaging modalities and anatomical structures: (1) EchoNet, (2) White Blood Cell (WBC), (3) HipXRay-Femur, and (4) HipXRay-Pelvis. EchoNet comprises echocardiograms with left ventricle line segment annotations at end-systole and end-diastole, which we converted to binary segmentation masks~\cite{ouyang2020video}. The WBC dataset features microscopy images of white blood cells paired with nuclei segmentation masks~\cite{Zheng2018}. The HipXRay dataset contains X-ray images annotated for both the femur and pelvis~\cite{gut2021xray}; we separated this into two independent tasks (HipXRay-Femur and HipXRay-Pelvis) using their respective structure labels as binary masks. For each dataset, we randomly partitioned the images into disjoint query and support pools using a 20:80 split, keeping this partition fixed across all experiments. To bound compute, we limited our evaluation to a random subset of 500 queries for datasets exceeding this size (EchoNet only).

\paragraph{In-context segmentation and image embedding models.} We used MultiverSeg~\cite{wong2025multiverseg} as the in-context segmentation model $f$ and a frozen DINOv3~\cite{simeoni2025dinov3} encoder as $g$ for all image embeddings. We converted images to grayscale for mask prediction with MultiverSeg, but used their RGB versions as inputs to DINOv3 for embedding generation. To optimize compute, we generated query and support embeddings once per dataset and reused them across all support sizes. Finally, we used FAISS~\cite{douze2025faiss} for efficient similarity searches of the image embeddings.

\paragraph{Support set sizes.} For both support set selection and failure detection experiments, we evaluated $K \in \{1, 2, 4, 8, 16, 32\}$ support examples, yielding $|K|=6$ distinct support set sizes.

\paragraph{In-context segmentation failure detection.} We constructed the labeled datasets for failure detection directly from the random sampling results of the support set selection experiment. By using sampled query–support set pairs $(x_q, \mathcal{S})$ across all independent trials and sizes, we generated $N \times |K| = 40 \times 6 = 240$ support sets per query. Because raw IoU scores vary based on task difficulty and structure size, we defined the binary failure threshold $\tau$ separately for each dataset. Specifically, we set $\tau$ to the dataset's median IoU across all pairs, which naturally balanced the binary failure label distribution. We report these dataset-specific thresholds in Table~\ref{tab:failure_metrics}.

We partitioned these labeled datasets into training and testing sets using a 75:25 query-level split, ensuring queries seen during training never appeared in the test set. We trained a separate failure classifier for each dataset using a four-layer transformer with eight attention heads. Models were trained for 20 epochs using cross-entropy loss and the Adam optimizer~\cite{kingma2014adam} (batch size 64, initial learning rate $10^{-3}$). All experiments utilized a fixed random seed. Finally, we quantified test-set performance using standard classification metrics: accuracy, F1-score, area under the receiver operating characteristic (AUROC), and area under the precision-recall curve (AUPRC).

\subsection{Support Set Selection}
\label{sec:results-selection}

\begin{figure}[t]
    \centering
    \includegraphics[width=\textwidth]{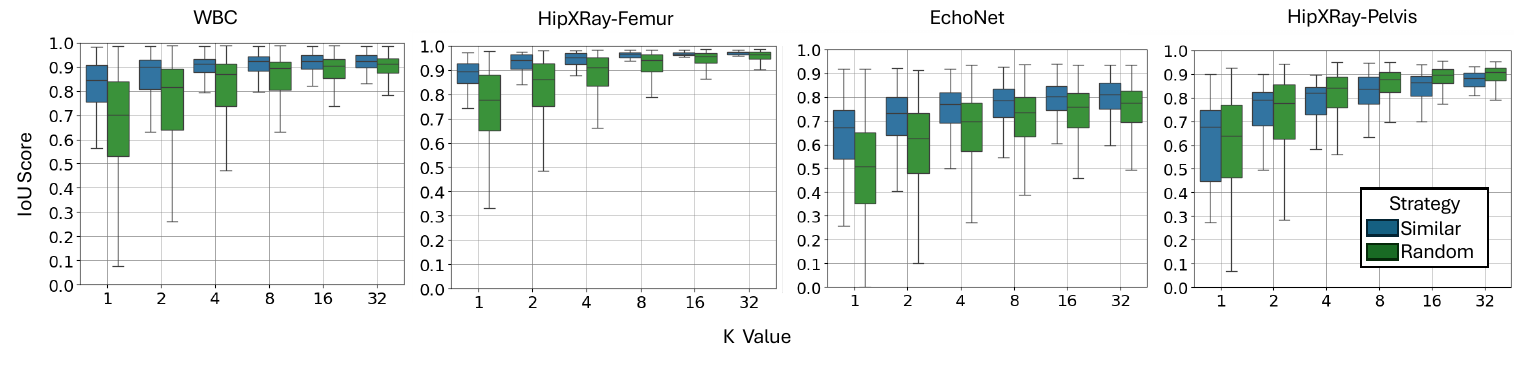}    
    \caption{\textbf{In-context segmentation performance for random and similarity-based sampling.} Box plots of IoU score distributions across four benchmarks (WBC, HipXRay-Femur, EchoNet, and HipXRay-Pelvis) as a function of support set size $K \in \{1, 2, 4, 8, 16, 32\}$ for random (green) and similarity-based (blue) selection. Performance improves with $K$ under both strategies. Similarity-based selection consistently matches or outperforms random sampling, yielding the largest gains at small $K$ where the support set is most constrained.}
    \label{fig:selection_boxplots}
\end{figure}

Figure~\ref{fig:selection_boxplots} presents the IoU score distributions for both sampling strategies across all four benchmarks and support set sizes. Two trends emerge consistently. First, under both strategies, performance improves logarithmically as the support set size $K$ increases. Specifically, median IoUs rise and interquartile ranges tighten. We observe the greatest gains between the smallest sizes ($K=1$ to $K=8$), with diminishing returns thereafter. Second, similarity-based selection consistently matches or exceeds the performance of random sampling at every support size, yielding both higher median IoUs and tighter interquartile ranges.

The advantage of similarity-based selection is most pronounced at small $K$, where the support set is most constrained and the choice of exemplars matters most. At $K=1$, selecting the single most similar reference raises the median IoU by 0.16 points on EchoNet and 0.14 points on WBC relative to a random draw. This approach also produces visibly narrower distributions, indicating higher typical performance and fewer low-quality outcomes. As $K$ grows, the two strategies converge. By $K=16$, the performance gap largely closes, as a sufficiently large random support set becomes increasingly likely to contain exemplars resembling the query regardless of how we assemble it.

The benchmarks also differ in absolute difficulty. HipXRay-Femur proves the easiest, with both strategies exceeding a 0.95 median IoU by $K=16$. Conversely, EchoNet remains the most challenging: the random sampling strategy plateaus around 0.76, while the similarity-based strategy plateaus near 0.80. Importantly, the benefit of similarity-based selection holds regardless of a dataset's overall difficulty. This suggests that informed support selection is a broadly applicable lever for in-context learning, rather than an artifact tied to a specific modality or anatomical structure.

\subsection{In-context Segmentation Failure Detection}
\label{sec:results-prediction}

\begin{table}[t]
    \centering
    \caption{\textbf{Evaluation metrics for in-context segmentation failure detection on the test set.} Overall classification metrics for the failure classifier across all four benchmarks, pooled across support set sizes. The IoU threshold ($\tau$) denotes the dataset-specific median IoU used to define failure. The AUROC exceeds chance (0.50) on every dataset, demonstrating that segmentation failure carries a strong signal recoverable from the query and support embeddings alone.}
    \label{tab:failure_metrics}
    \begin{tabular}{|c|c|c|c|c|c|}
    \hline
    Dataset & $\tau$ & Accuracy & AUPRC & AUROC & F1 Score \\
    \hline
    EchoNet & 0.70 & 0.73 & 0.79 & 0.80 & 0.73\\
    WBC  &  0.87 & 0.62 & 0.68 & 0.69 & 0.62\\
    HipXRay-Femur & 0.92 & 0.71 & 0.77 & 0.78 & 0.71 \\
    HipXRay-Pelvis & 0.86 & 0.66 & 0.67 & 0.71 & 0.66 \\
    \hline
\end{tabular}

\end{table}

\begin{figure}[t]
    \centering
    \includegraphics[width=\linewidth]{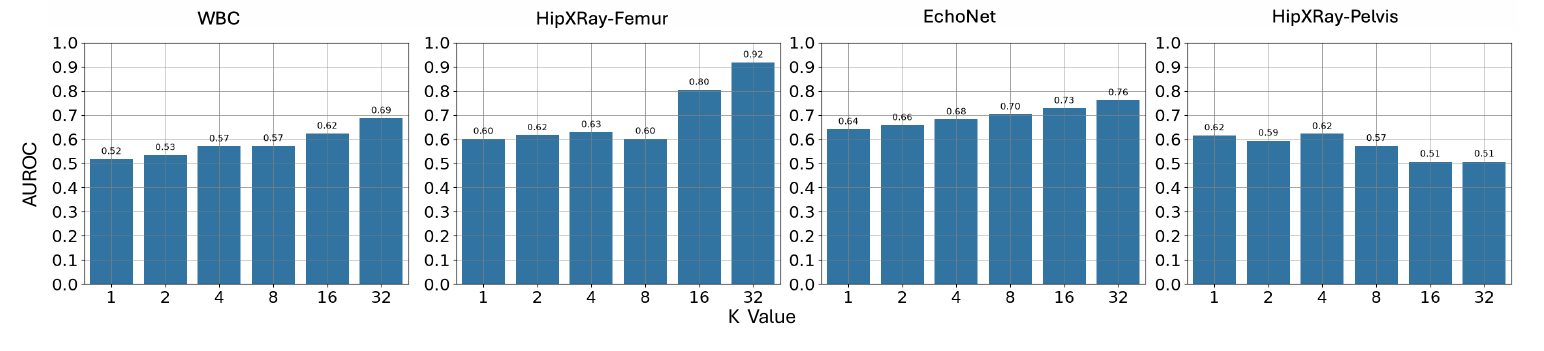}
    \caption{\textbf{Failure detection performance as a function of support set size.} Test-set AUROC of the failure classifier at each support set size $K \in \{1, 2, 4, 8, 16, 32\}$ across the four benchmarks (WBC, HipXRay-Femur, EchoNet, and HipXRay-Pelvis). On three of the four datasets, failure becomes more predictable as the support set grows. HipXRay-Pelvis is the exception, where the AUROC declines toward chance at large $K$.}
    \label{fig:auroc_k}
\end{figure}

Table~\ref{tab:failure_metrics} reports the overall failure detection performance on the held-out test set queries for each benchmark. The classifier shows moderate performance across all four datasets, with AUROC scores consistently exceeding chance (0.50). Failure detection performance is highest on EchoNet (AUPRC: 0.79; AUROC: 0.80), and lowest on WBC and HipXRay-Pelvis (AUPRC: 0.68 and 0.67; AUROC: 0.69 and 0.71, respectively).

Figure~\ref{fig:auroc_k} breaks this performance down by support set size. On three of the four benchmarks (EchoNet, WBC, and HipXRay-Femur), the AUROC rises with $K$, indicating that segmentation success or failure becomes more predictable as the support set grows. This effect is most pronounced on HipXRay-Femur: the AUROC sits at 0.60 and 0.62 for small support sets ($K=1$ and $2$, respectively), but climbs sharply to 0.80 and 0.92 at $K=16$ and $32$. HipXRay-Pelvis is the exception; its AUROC remains near chance and actually declines at large $K$.

\section{Discussion}
\label{sec:discussion}
In this work, we investigated support set construction for in-context medical image segmentation along two axes: how its composition can be improved through informed selection, and how the resulting segmentation quality can be anticipated before it is trusted. Our selection results demonstrate that similarity-based sampling consistently matches or outperforms random sampling across every support size and dataset. Notably, the largest gains occur at small $K$, precisely the low-annotation regime that makes in-context segmentation so attractive for clinical deployment. Furthermore, our failure detection results establish that segmentation quality is predictable using query and support embeddings alone, with performance exceeding chance across all four benchmarks. Together, these findings suggest a highly practical deployment pattern: first retrieve support sets by similarity, then screen by a lightweight detector that flags cases where even the best available exemplars are unlikely to yield a reliable result.

\subsection{Limitations}
We evaluated only a single segmentation model (MultiverSeg) and a single image encoder (DINOv3), leaving it an open question whether these benefits generalize to other architectures or medical-specific encoders. Furthermore, both our selection and failure detection mechanisms rely exclusively on image embeddings rather than masks, making them blind to the specific target structure. Finally, we derived our failure labels from a single median-IoU threshold, and we evaluated our benchmarks using a single fixed data split. Future work must investigate model behavior under task-specific thresholds and out-of-distribution shifts.

\begin{credits}
\subsubsection{\ackname} The authors declare no funding source associated with this work. 

\subsubsection{\discintname}
The authors declare no competing interests. 
\end{credits}

%
%
%
\bibliographystyle{splncs04}
\bibliography{references}
%





\end{document}